\documentclass[9.5pt,journal,final,finalsubmission,twocolumn]{IEEEtran}

\pdfoutput=1
\usepackage{epsf,amsmath,amssymb,graphicx,epsfig,subfigure}
\usepackage{hyperref}
\usepackage[lined, algonl]{algorithm2e}

\graphicspath{ {images/} }

\title{Agile, Antifragile, Artificial--Intelligence--Enabled Command and Control}

\author{%
\begin{tabular}{c}{\hspace{-0.5cm}}FLTLT Jacob Simpson \\ {\hspace{-0.5cm}}University of New South Wales, Canberra \\{\hspace{-0.5cm}} jacob.simpson@student.unsw.edu.au \end{tabular} \and
\begin{tabular}{c}{\hspace{0.5cm}} Dr Rudolph Oosthuizen \\ {\hspace{0.5cm}}University of Pretoria, South Africa \\{\hspace{0.5cm}} rudolph.oosthuizen@up.ac.za \end{tabular} \and
\begin{tabular}{c} Dr Sondoss El Sawah \\ University of New South Wales, Canberra \\ s.elsawah@adfa.edu.au \end{tabular} \and
\begin{tabular}{c} Dr Hussein Abbass \\ University of New South Wales, Canberra \\ h.abbass@unsw.edu.au \end{tabular} }

\begin{document}

\maketitle
\setcounter{page}{1}
\begin{abstract}

Artificial Intelligence (AI) is rapidly becoming integrated into military Command and Control (C2) systems as a strategic priority for many defence forces. The successful implementation of AI is promising to herald a significant leap in C2 agility through automation. However, realistic expectations need to be set on what AI can achieve in the foreseeable future. This paper will argue that AI could lead to a fragility trap, whereby the delegation of C2 functions to an AI could increase the fragility of C2, resulting in catastrophic strategic failures. This calls for a new framework for AI in C2 to avoid this trap. We will argue that \textquoteleft antifragility\textquoteright \ along with agility should form the core design principles for AI-enabled C2 systems. This duality is termed Agile, Antifragile, AI-Enabled Command and Control (A3IC2). An A3IC2 system continuously improves its capacity to perform in the face of shocks and surprises through overcompensation from feedback during the C2 decision-making cycle. An A3IC2 system will not only be able to survive within a complex operational environment, it will also thrive, benefiting from the inevitable shocks and volatility of war.

\end{abstract}

\section{Introduction}

The integration of Artificial Intelligence (AI) into military Command and Control (C2) is seen by many as a crucial element to establish a competitive edge for a military force~\cite{ADF_C2:2018,UK_DOD_C2:2017,Kania:2017}. Expectations of what AI can achieve on the battlefield are high, with some declaring it the next \textquoteleft revolution in military affairs\textquoteright~\cite{Wallace:2018}. AI is expected to automate complex functions within C2, leading to the concept of the \textquoteleft battlefield singularity\textquoteright, whereby the increase in pace of operations from the automation of the decision-making cycle results in human cognition being unable to keep up with the machine-speed in which decisions are made~\cite{Kania:2017}. Within this vision for the future battlefield, the human is seen to be a weak link in C2 systems~\cite{Mousavi:2020}.

This paper argues that the integration of AI could have unintended consequences on the performance of C2 systems seeking machine-speed decision-making; that strategically, a system which has reached a \textquoteleft battlefield singularity\textquoteright \ is fundamentally fragile. The rapid development of AI and its obvious revolutionary / disruptive implications for C2 systems has been largely led by a focus on the degree of \textquoteleft responsiveness\textquoteright \ to an opponent during war, not on how such technology may impact C2 system performance holistically. Two assumptions are made in the literature: first, it is assumed that AI will further the goal of improving agility through optimisation of the parts of a system; and second, that future AI-enabled C2 systems will improve with as little human input as possible, due to a sophisticated AI capable of wartime decisions, even at the strategic level~\cite{Li:2018, Wang:2020}. Both of these assumptions are misplaced, as AI brings unique qualities that could add to the fragility of C2 systems.

Traditionally, C2 systems have been argued to benefit from a strategy that focuses on maximising agility within a complex competitive environment~\cite{Alberts:2011, Oosthuizen_and_Pretorius:2014, Jensen:2012, Berggren:2014}. David Alberts has exemplified this strategy with the \textquoteleft Agile C2\textquoteright \ concept, which states that in order for a C2 system to be effective, it must be able to successfully cope, exploit, and effect change within a complex environment. C2 effectiveness is achieved through the interaction of system elements such as adaptability, responsiveness, flexibility, versatility, innovativeness, and resilience~\cite{Alberts:2011}. However, the acceptance of the Agile C2 model has led the majority of military C2 doctrine and literature to incorporate AI technology as a means to increase the \textit{responsiveness} of C2 decision-making alone~\cite{ADF_C2:2018, UK_DOD_C2:2017, Wang:2020, Kania:2017, Mousavi:2020}, while less attention was paid to the mere fact that a C2 system needs to be responsive to meet strategic interests. Here lies the core of the problem, whether or not an AI that improves responsiveness will be able to do so while understanding the consequences of decisions on strategic and grand-strategic objectives across multiple-domains. We argue that, despite AI sophistication, predictions within an operational environment are fundamentally fragile due to the vulnerability of AI-Enabled Systems to Black Swan events with strategic consequences~\cite{Wallace:2018}. The optimisation qualities of AI, coupled with diminished human responsibilities, could become a \textquoteleft fragilising\textquoteright \ process that hinders C2 agility.

To negate some of the issues identified above that could lead to fragility within AI-enabled C2 systems, a new design principle that enhances a system\textquoteright s ability to improve itself from volatility, known as \textquoteleft antifragility\textquoteright, is required~\cite{Taleb:2012, Taleb:2020}. Properly-designed AI could enable the development of an antifragile system by accumulating appropriately encountered and learnt experiences in a system-level memory, but it may also encourage the over-optimisation of the C2 decision-making cycle. This could result in a system being unable to recognise and interpret unexpected events, but still recommending decisions rapidly, leading to an escalation in negative risks. The integration of AI thus supports the development of a new model, extending the concept of Agile C2 with the inclusion of antifragility. This will be termed Agile, Antifragile, AI-Enabled Command and Control (A3IC2), and is the amalgamation of Agile C2, Antifragility theory, and AI for C2, building upon the previous models developed by Boyd, Brehmer and Alberts~\cite{Brehmer:2005, Alberts:2011}.

In order to explore A3IC2, the paper is structured as follows. A literature review is presented in Section~\ref{sec:literature_review} to distinguish A3IC2 concept from the others that have preceded it. The rationale for expecting AI to lead to fragility is then presented in Section~\ref{sec:AIAntifragility} followed by an argument of the reasons that antifragility will enable effective use of AI in a C2 system in Section~\ref{sec:Antifragility}. The proposed A3IC2 functional model is discussed in Section~\ref{sec:A3IC2}. Conclusions are then drawn in Section~\ref{sec:conclusion}

\section{Literature Review}\label{sec:literature_review}

\subsection{Command and Control}

The definition of Military C2, for the purposes of this paper, is the theatre level function responsible for the appropriate \textit{allocation of forces} in order to achieve \textit{military objectives}. Military doctrine widely defines it as the \textquoteleft process and means for the exercise of authority over, and lawful direction of, assigned forces\textquoteright~\cite{ADF_C2:2018, UK_DOD_C2:2017, Marines_C2:2018}. This is distinct from other systems described as C2 at the tactical level, such as C2 for individual vehicles or small units. 

Military C2 is inseparable from strategic decision-making. It consists of a hierarchical organisation, with the commanders intent, derived from the strategic objectives of the nation that they are defending, supplying the direction for decisions and actions undertaken by subordinates~\cite{Brehmer:2005}. One of the highest priorities for C2 is maintaining situational awareness of the environment and responding (or not) appropriately with military action to achieve strategic objectives. Not only does C2 have to conduct battle effectively, but it also must know when to transition from Operations Other Than War (OOTW) to battle~\cite{Albino:2016} and vice-versa. An appropriate abstraction (or model) of military C2, therefore, needs to acknowledge the full spectrum of conflict; from war to OOTW ~\cite{Marines_C2:2018}. It must take into account the dynamic complexity of the \textquoteleft operational environment\textquoteright \ in which the C2 system is a part; from the tactical to the strategic level and the effects it generates on the grand-strategic level. In short, effective C2 is not merely one that can win battles, it must also know when instigating battle is a proportionate response~\cite{UK_DOD_C2:2017, Marines_C2:2018, Albino:2016}. Moreover, it needs to understand the implications of its actions on the grand-strategic level; that is, the whole-of-government objectives.

C2, as a system, operates within an environment that is nonlinear and complex. It is classified as a \textquoteleft sociotechnical\textquoteright \ system, a mix of technological and \textquoteleft social\textquoteright \ or human elements that interact with one another and a wider complex environment~\cite{Walker_etal:2008}. A C2 system exhibits dynamic, emergent behaviour with many unintended or unpredictable consequences. This is not only due to the fact that these systems rely on humans to make sense of the complex environment and to develop the plans to solve problems, but because it is also a technical system, with situational awareness reliant on digital systems and sensors to pass information that may not accurately represent the operational environment~\cite{Oosthuizen_and_Pretorius:2014, Walker_etal:2008, Jensen:2012, Wallace:2018}. The missions or objectives that a C2 system must accomplish is wholly dependent on unanticipated real-world events, such as wars, environmental disasters, and other miscellaneous OOTW. This occurs in multiple domains (physical and non-physical) and all under the effect of friction. From a systems thinking perspective, the C2 operational environment is truly \textquoteleft hyper-complex\textquoteright~\cite{Grosser:2017, Albino:2016}.

Military C2, therefore, has a very difficult task, in that it must make highly consequential decisions in a complex environment with guaranteed second and third-order strategic effects that are near impossible to predict or reverse~\cite{Avila:2016, Wallace:2018}. This has long been understood by military strategists, and has been traditionally managed through mental models or heuristics for guidance on how to understand and respond to the complexity of war. These mental models are now cemented in the strategic studies discipline and modern military doctrine~\cite{Wallace:2018}. C2 is an essential means of achieving strategic success in war, which is defined as \textquoteleft determining a method to cause the collapse of the enemy\textquoteright s organisation due to helplessness or confusion\textquoteright~\cite{Albino:2016}. The mental models associated with guiding this outcome are (by necessity) highly abstracted, reflecting an understanding of complexity; that strategy is more of an art than being a science. Clausewitz and his concept of \textquoteleft friction\textquoteright, describes the difficulties of operating within this complexity, with its habit of destroying all carefully orchestrated plans, resulting in the observation that \textquoteleft everything is very simple in war, but the simplest thing is difficult\textquoteright~\cite{Clausewitz:1874}. The heuristics of strategy have progressed since Clausewitz due to the significant advancements in information theory, AI, systems thinking and cybernetics. Mental models on war continue to evolve from technology, but the core nature of war has not. Its foundation in politics requires that it is an activity inseparable from the human element~\cite{Wallace:2018, Marines_C2:2018}. Transforming these mental models into concrete metrics to guide an AI is a non-trivial, possibly infeasible, task. These mental models work on a holistic understanding of the context, the commander\textquoteright s intent, and the grand-strategic consequences that a decision could produce.

Science, technology and information theory have had a significant impact on strategic and C2 concepts~\cite{Osinga:2007}. Colonel John Boyd, as a student of cybernetics and strategy, has built upon the work of both disciplines to create one of the most influential functional models in the strategic studies field, the Observe Orient Decide Act (OODA) Loop. The OODA Loop is a model detailing a theory for \textquoteleft winning and losing\textquoteright, broadly describing how one can manage a competitive environment and survive~\cite{Osinga:2007}. For an effective and survivable C2, Boyd argued that a system must be able to adapt to its environment faster than the enemy. The Orient step represents making the \textquoteleft right decision\textquoteright \ based on observations, analysis and mental models, but if all else is equal between both opponents, whoever loops through each step faster will win~\cite{Osinga:2007}. Therefore, the C2 system that drives conflict at a rate more rapid than the adversary can respond, will cause a \textquoteleft fatal destabilization\textquoteright, thus achieving victory~\cite{Wallace:2018}. It is from the development of the OODA Loop theory that the systems thinking C2 literature continues its study of what makes a superior C2 system; a multi-disciplinary area combining the systems thinking approach and strategic studies~\cite{Brehmer:2005, Avila:2016, Osinga:2007}. There is a broad agreement within the literature that the complexity of war necessitates that a C2 system be dynamic or agile, allowing one to both achieve victory and avoid system failure~\cite{Alberts:2011, Jensen:2012, Berggren:2014, Oosthuizen_and_Pretorius:2014, Osinga:2007, Wallace:2018}.

\begin{figure*}[ht]
\centering
\includegraphics[scale=0.90]{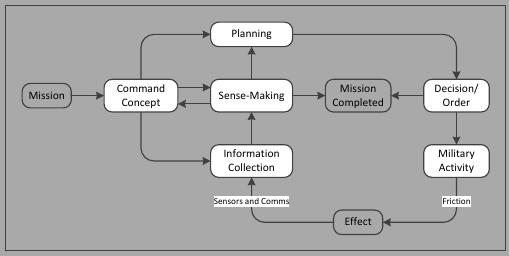}
\caption{Dynamic OODA Loop~\cite{Brehmer:2005}}\label{fig1}
\end{figure*}

However, although the OODA Loop is sound as a theory of winning and losing, it is not as sufficient a model for implementing agility in a C2 system as it neglects specific functions, such as \textquoteleft command concepts, planning, exit criteria or system delays\textquoteright, resulting in a model that overly accentuates speed as the aim~\cite{Brehmer:2005, Oosthuizen_and_Pretorius:2014, Avila:2016}. In order to incorporate the OODA Loop as a better model for C2, Brehmer developed the Dynamic OODA Loop (DOODA Loop). Brehmer argues that specific details are needed, such as the delays throughout the decision making process, in order for the model to be descriptively adequate in the C2 context~\cite{Brehmer:2005}. The DOODA loop, seen in Figure~\ref{fig1}, therefore allows for commanders and staff to actually understand each function of the C2 process. It illustrates what needs to be achieved in order to improve agility and decision-making, by making each C2 function explicit ~\cite{Brehmer:2005}. For this reason, the DOODA loop model will be used as the basis for the A3IC2 functional model later within this paper.

From the above discussion, one concept is clear: C2 and its measures of performance are inseparable from the strategic context in which the system is operating. The dynamics within a C2 system do not take place in a vacuum; the final outcome of a C2 system is the affect of \textit{control}, or ability to make effective decisions within the hyper-complex environment in the command of military force, \textit{in order to survive and win}. If a highly sophisticated, efficient, and responsive AI-enabled C2 system is unable to track the complexity of the operational environment, the effects generated, and their consequences on a grand-strategic level, the C2 system will not survive under the volatility of high-intensity warfare.

\subsection{C2 System Definitions}

Within the literature, the descriptions of C2 system types are problematically similar, resulting in considerable overlap with the definitions of agility, adaptation, robustness and resilience depending on the situation or context~\cite{Berggren:2014, Bruijn:2018, Alberts:2011}. However, there are two broad fundamental methods for survival being described that all C2 system types share at least one aspect of:
\begin{enumerate}
    \item Strength in retaining form (the ability to survive volatility without change)
    \item Changing form to retain strength (the ability to survive volatility through change)
\end{enumerate} 
Both methods for survival can be effective depending on the circumstances; therefore, a useful C2 functional model must include both. The C2 literature broadly understands this, and has sought to combine various definitions in functional models to reconcile both methods~\cite{Berggren:2014}. The concept of \textquoteleft Agile C2\textquoteright \ incorporates resilience and robustness into its definition, departing from the usual understanding of agility as simply meaning \textquoteleft swiftness\textquoteright \ in changing form. Alberts defines Agile C2 as \textquoteleft the ability to successfully effect, cope with, and/or exploit changes in circumstances\textquoteright~\cite{Alberts:2011}. This definition has six dimensions required to achieve that end~\cite{Alberts:2011,Carreno:2020}: responsiveness, flexibility, adaptability, versatility/robustness, innovatiness, and resilience.

The fusion of all these elements is expected to minimise the probability of events associated with adverse impacts, and maximise events that offer opportunities. These elements also work to minimise the cost or maximise the gains should the event actually occur~\cite{Alberts:2011}. It is stressed that a single-objective optimisation does not equate to agility; instead, it reflects an imbalance towards responsiveness over flexibility and resilience. A system that is effectively agile will not necessarily be efficient when its optimisation relies on a single-objective, even when this single objective is a pre-determined weighted sum of different objectives. We acknowledge, however, that optimisation is a mathematical concept that could be adapted to achieve any objective. If the aim is to balance responsiveness, speed, flexibility, and resilience, multi-objective optimisation is a branch of optimisation theory that could handle this problem mathematically and optimise conflicting objectives simultaneously.

The objective of Agile C2 to minimise adverse impacts and maximise opportunities resembles Nassim Taleb\textquoteright s idea of a \textquoteleft convex\textquoteright \ system; a beneficial response to volatility, otherwise known as antifragility~\cite{Taleb:2012}. Agility and antifragility have many similarities. Both agility and antifragility share a view of risk that seeks to both reduce the negative impact of Black Swan events (catastrophic, low probability occurrences) and avoid the complacency that underestimates their likelihood within an organisation~\cite{Alberts:2011,Taleb:2012}. Other similarities are seen with the listed qualities that an organisation should avoid if it is to be an antifragile organisation, such as limiting the use of single-objective optimisation, specialisation, forecasting, standardisation, and micromanagement~\cite{Blecic:2019, Taleb:2012, Alberts:2011}.

Like Agile C2, the antifragile organisation focuses on policy and structures that maximise freedom of action (flexibility). It discourages optimisation, lack of diversity, risk intolerance, and crucially, an unrealistic simplified model of reality~\cite{Alberts:2011,Blecic:2019}. However, the crucial difference between antifragility and Agile C2, is \textquoteleft the \emph{purposeful} implementation of induced small stressors\textquoteright \ or \textquoteleft non-monotonicity\textquoteright \ into a system for the purposes of learning and overcompensation~\cite{Kennon:2015, Johnson_Gheorghe:2013, Taleb:2012}. This is the key variable between an antifragile system and an agile or a resilient system. Antifragile systems actively seek to inject volatility within its own system in order to expose fragility. The differences between the two concepts are complementary, and it will be argued that when both are combined, can produce a robust functional model for an AI-enabled C2 system.

\subsection{Antifragility and C2}

\begin{table*}
\caption{Elements of the A3IC2 System~\cite{Alberts:2011, Taleb:2012, Johnson_Gheorghe:2013, Bruijn:2018}}\label{tab1}
\begin{tabular}{p{0.2\columnwidth}p{0.5\columnwidth}p{0.4\columnwidth}p{0.5\columnwidth}p{0.2\columnwidth}}
\hline
\hline
\textbf{System Elements} & \textbf{Definition}  & \textbf{Effect on performance from change?}  & \textbf{What happens to the system upon change?} & \textbf{Reactive or Proactive?}\\ \hline\hline

Adaptability    & The ability to change ones own system, organisation and/or structure to become better suited for the challenge.     & Maintains a minimum level of performance and returns to normal functioning over time. A similar future shock will have less effect. & System maintains a minimum level of performance and returns to normal functioning in an acceptable amount of time via system change.    & Reactive\\ \hline

Responsiveness  & A Systems ability to respond, proactive or reactive, to a change in circumstances, be it a stress or opportunity.   & When a change occurs, the performance drop from baseline  is mitigated through rapid response.  & System will maintain a level of performance above the minimum acceptable level. Will return to normal functioning at a more rapid rate. & Reactive/Proactive\\\hline 

Flexibility
(Optionality)   & Ability to adapt a response to change in more than one way to accomplish a task. System is convex in design, enabling positive exploitation from shocks, while minimising any downside (\textquoteleft barbell strategy\textquoteright).    & Minimises performance loss from volatility through selection of an appropriate response. Is not forced to take an inefficient response. Will maximise gains in performance. & If a negative shock, system will have a minimal performance loss. If a positive shock, the system has the options available to rapidly improve through exploitation.    & Proactive\\ \hline

 Innovativeness
(overcompensation)  & Permits and entity to generate or develop a new tactic or way of accomplishing something. Risk taking and invention incorporates the antifragile qualities of overcompensation and small scale experimentation. & System will improve in performance through learning from feedback to adapt in order to overcompensate.  & Will improve in agility and/or resilience based on the specifics of the overcompensatory action. Will survive harsher shocks beyond those previously experienced.   & Reactive/Proactive\\ \hline

Memory/Feedback & Ability of the C2 system to collect, store and maintain experiences and lessons from the operational environment.   & Performance is improved overtime through the systems ability to store information for adaptation.   & System learns effective methods for historical problems through abstraction, improving adaptability and resilience.   &   Reactive\\ \hline

Resilience  & Ability to withstand interruption/degradation and return to normal operational capacity. A system that absorbs the impacts of stress or shocks and reorganises itself after. &    Maintains a minimum level of performance and returns to normal functioning. & System maintains a minimum level of performance and returns to normal functioning in an acceptable amount of time.  & Reactive\\\hline 

Versatility (Robustness)    & An acceptable level of performance or effectiveness in accomplishing new or significantly altered task or mission. Is reliable to expected and unexpected inputs.   & Systems behaviour shows a satisfactory response to seemingly extreme conditions and is insensitive to change.   & Shows no significant changes to randomness due a system mechanism that provides strong balancing loops.     & Reactive \\ \hline\hline

\end{tabular}
\end{table*}

Antifragility is a system quality, or feature, that enables it to not only be robust and resilient to sudden shocks and stressors, but to also learn from these stressors to improve itself next time it encounters them~\cite{Taleb:2012, Baxter:2017}. Antifagility is the true opposite of fragility, as neither the definitions of robustness or resilience \textquoteleft imply gains in strength from shocks\textquoteright~\cite{Albino:2016, Taleb:2012}. Taleb states that an antifragile system \textquoteleft has a mechanism by which it regenerates itself continuously through using, rather than suffering from, random events, unpredictable shocks, stresses and volatility\textquoteright~\cite{Taleb:2012}. This follows that antifragility is \textquoteleft not possible if there is no mechanism for feedback and for memory\textquoteright~\cite{Baxter:2017}. Therefore, in order to manoeuvre a system towards antifragile system dynamics, it must enable methods to learn from shocks to its system (feedback) and improve its operations from this memory (orientation). It is vital to emphasise that the feedback could be internal and self-generated using an internally designed measures of performance and effects, together with a role-play of scenarios using internal simulations of an external environment. As a concept, antifragility has the following five dimensions\cite{Taleb:2012, Derbyshire:2013, Kennon:2015}:
\begin{enumerate}
    \item An ability to learn from shocks and harm: The system has the capability to store its memory and experience from the feedback it receives. 
    \item The use of overcompensation for system improvement: Once feedback is received, the system will improve itself \textit{beyond} what is required to manage a similar shock in the future.
    \item Redundancy: The system will develop multiple levels of redundancy as a result from overcompensating from shocks. 
    \item Convexity and optionality (\textquoteleft barbell strategy\textquoteright): The system will structure itself in a way to maximise potential gains but minimise potential losses. In other words, the system will be robust but ready to exploit gains.
    \item Small-scale experimentation: Risk taking in order to achieve significant performance gains at the expense of small failures. Induced small stressors on the system to ensure non-monotonicity. 
\end{enumerate}

The three features that separate the Agile systems from antifragile ones are the focus on overcompensation, the purposeful inducing of system stress, and memory/feedback from volatility. The antifragile system will improve itself to not just be able to compensate for a similar stressor in the future, but for an even harsher shock than what was experienced~\cite{Taleb:2012}. Volatility is therefore highly desirable as it allows the system to gather information and future-proof itself through learning from as wide variety of input as possible. This generates the data needed for an overcompensatory adaptation to the system to manage the shock. In fact, an antifragile system will purposely attempt \textquoteleft risk-managed experimentation\textquoteright \ in order to create the volatility needed to overcompensate. Taleb explicitly states that this includes the risks from Black Swans; events that have a high improbability and extreme impact~\cite{Kennon:2015, Derbyshire:2013, Alberts:2011}. Black Swans are of high value to an antifragile system due to the rare information gained for strengthening the system, as long as they are initially survivable~\cite{Taleb:2012}, hence, the importance of resilience and robustness. The antifragile system is designed to be as survivable as possible against chaos as an ontological reality, unable to be removed or predicted within complex environments~\cite{Derbyshire:2013, Taleb:2012}. 

Alberts~\cite{Alberts:2011} discusses the conceptual model of agility, with the \textquoteleft circumstances space\textquoteright \ representing the systems level of performance depending on various external and internal changes. From the Agile C2 perspective, an antifragile system \textit{explores} the circumstance space in order to understand as many \textquoteleft areas of acceptable performance\textquoteright \ from as many generated circumstances as possible. Volatility and feedback allows for this exploration. The effective use of feedback/memory and a focus on experimentation through volatility in order to overcompensate, thus enables the Agile C2 system to increasingly understand its \textquoteleft model of self\textquoteright \ through exploration, improving its agility through a greater \textquoteleft variety of circumstances that an entity can recognize and successfully respond to\textquoteright~\cite{Alberts:2011}.  Moreover, the system develops a better understanding of the environment, the context within which shocks could be anticipated, and the environmental constraints that shape environmental stressors. Lessons learnt can take several forms, such as validated models of the operational environment, AI mathematical functions representing the environment, and the storage of other human/machine generated data. This information would be updated with new information from each shock, allowing for the C2 system to improve in effectiveness over time.

By now, it should be clear that an antifragile system does not preclude agility as a favourable characteristic within a system; antifragility is an additional trait - not a substitute~\cite{Taleb:2012, OReilly:2019}. In his definition of antifragility, Taleb splits agility away from the same spectrum as fragility, resilience and antifragility. For clarity in the A3IC2 construct, this will be continued. Seen below in Figure~\ref{fig2} is the agile and antifragile spectrum. The definition of each is divided into \textquoteleft systems that survive from volatility\textquoteright \ and the \textquoteleft ability for a system to enact change in order to survive\textquoteright. This neatly encapsulates the definitions described above in the system dynamics literature~\cite{Johnson_Gheorghe:2013}. For example, one cannot be resilient and return to normal level of performance upon shock without the ability of the system to recover and/or adapt in order to do so. Immutability is also fragile, as all systems function from the property of impermanence; without change a system will eventually fail~\cite{Rosenthal:2017}. Agility is an enabler for antifragility, as effective overcompensation on feedback requires an agile organisation; and the reverse is true, Agile C2 requires overcompensation to proactively innovate and build resilience from changes in the operational environment. 

\begin{figure}[ht]
\centering
\includegraphics[scale=0.30]{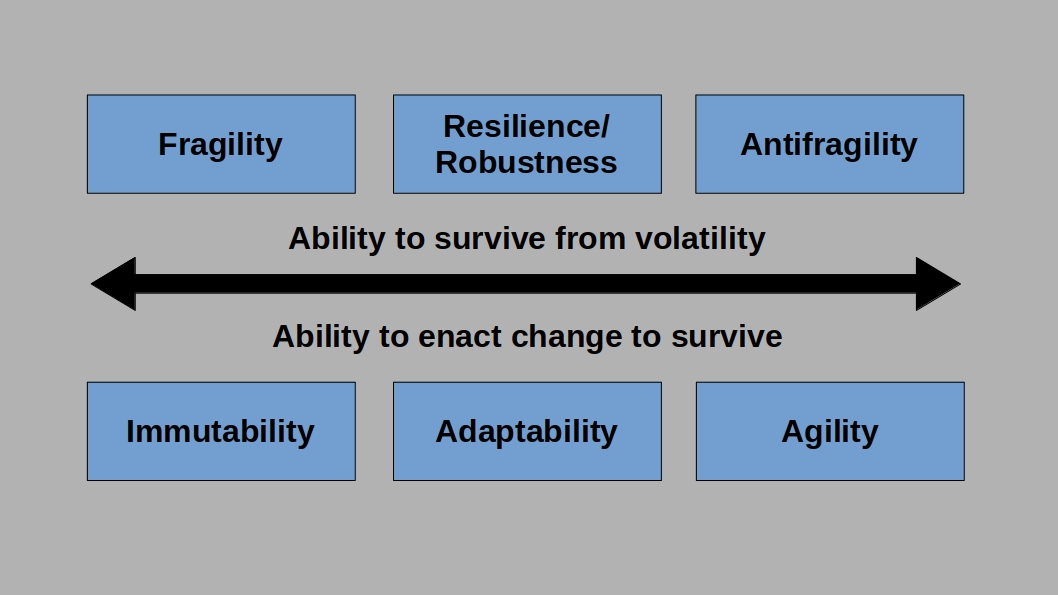}
\caption{Agile and Antifragile Spectrum~\cite{Alberts:2011, Taleb:2012}}\label{fig2}
\end{figure}

The benefits of combining agility with antifragility result in a much greater response to shocks when compared to resilient and robust systems~\cite{Bruijn:2018}. Taleb states that \textit{fragility} is mathematically \textquoteleft defined as an accelerating sensitivity to harmful stresses: this response plots as a concave curve and mathematically culminates in more harm than benefit from random events\textquoteright. A fragile system will collapse under extreme volatility, as it has no properties to negate a concave response. This follows that the dynamics of antifragility produce \textquoteleft a convex response that leads to more benefit than harm \textquoteright~\cite{Taleb:2012}. A resilient or robust system is therefore in the middle of the spectrum between fragility and antifragility. A robust and/or resilient system has neither much to gain nor lose from volatility. Antifragility has elements in place that allow it to not only return to normal functionality after a shock, but learn from stressors in order to overcompensate. Therefore, to obtain an antifragile and Agile C2 system, it requires the following elements listed in Table~\ref{tab1}.

As seen in Table~\ref{tab1}, the mix enables the strength of both approaches. The bottom three rows are antifragile elements, the first three rows are Agile C2 elements, and the middle row is a necessity for both. The seeking of innovative solutions to remove fragility and improve agility is required by both for overcompensation. Memory/feedback, optionality, and additions to innovativeness, are the new elements separating Agile C2 from A3IC2. How a C2 system can practically develop these elements requires an intersection of AI, Chaos engineering, and specific organisational strategies; the subject of the next section.

\section{Artificial Intelligence and Engineering Antifragile C2 Systems}\label{sec:AIAntifragility} 

Implementing antifragility within a C2 system requires the exploitation and accumulation of feedback regarding system performance; most readily achieved with the collection of data as a permanent method for retaining memory and learning within a system. This allows for the creation of antifragile feedback loops enabling the use of overcompensation~\cite{Johnson_Gheorghe:2013, Bruijn:2018}.  Jones~\cite{Jones:2014} describes an antifragile machine as one that can adapt to an unexpected environment due to its script becoming more complicated over time from the process of making decisions, taking action, and then observing the results. This machine must \textit{learn} from its environment and adapt to changes that were \textquoteleft not preconceived at design\textquoteright~\cite{Jones:2014}. In other words, to be truly antifragile, the scenarios that the system faces must be new, but also familiar enough to be generalised or abstracted from previous experiences, creating new knowledge. This process of a machine updating its internal states from its experience through interaction with the environment and/or sensed data is known as \textquoteleft machine learning\textquoteright \ (ML), a branch of AI. This technology is thus the basis by which antifragile dynamics can be achieved within a system~\cite{Jones:2014}.

A consensus within the literature on the definition of AI has not yet been reached, but for the purposes of this paper, AI is defined as \textquoteleft algorithms to provide computers with cognitive skills and competencies for sense-making and decision-making\textquoteright~\cite{Abbass:2019}. The methods for building AI systems vary. The traditional method is through \textquoteleft expert systems\textquoteright \ or \textquoteleft handcrafted knowledge\textquoteright, whereby algorithms are created through manually coding the solution with consultation from experts~\cite{Allen:2020, Antony:2016}. However, these systems are usually very brittle to a constantly changing environment due to the models being updated by hand. ML offers a substitute to update a system\textquoteright s knowledge, either from data that the system receives directly or through interaction with the environment. Advanced ML models such as deep learning rely on large datasets and specialised algorithms to learn specific patterns within structured (tabled) and unstructured (pictures, documents) data; allowing for the creation of a sophisticated mathematical representation/model of a system. Such a model could be used for making predictions on new data, or taking actions in previously unseen context. AI models can perform much more accurately against a complex environment, due to the multi-dimensional patterns within datasets collected from observations of the environment itself~\cite{Allen:2020}. AI promises to reduce the many limitations of human decision-making, such as  attention-focus, limited memory, recall, and information processing~\cite{Sterman:2000}.

\begin{figure}[ht]
\centering
\includegraphics[scale=0.45]{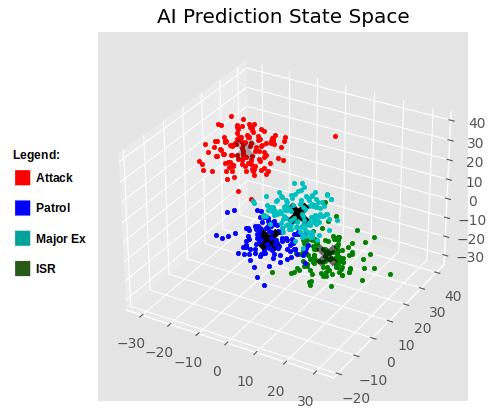}
\caption{Highly Simplified State Space with Topology Formed from ML Clustering Algorithm}\label{fig3}
\end{figure}

ML methods attempt to functionally approximate a high-dimensional topology within a space~\cite{Wallace:2018}. The system that the data is derived from provides the topology via sensors, and the ML algorithm attempts to learn this topology through training and then validating its performance (ie accuracy). When a new data point is presented to a trained AI, it gets placed within this same configuration space, and depending on the approximations that the algorithm has formed, it will make a prediction on the new data point. As an example, Figure~\ref{fig3} is a low-dimensional result of a ML classification algorithm. It has four labels representing a prediction for the enemies current behaviour, each designated by an AI designer based on previous understanding of the data. When a new data point is received and evaluated within this state space; the data point may get assigned to the closest cluster. If the euclidean distance from the data point is closest to the red cluster, then the AI would then output a \textquoteleft probable attack\textquoteright \ as a prediction, possibly with a likelihood derived from the distance to the red dots compared to distances to other clusters.

The argument for AI being an enabling tool for an Agile C2 system is, therefore, fundamentally reducible to the utility in forming these adaptive complex mathematical functions to model a dynamic and changing environment. It is argued that these models will provide higher accuracy than humans for most C2 tasks and rapid and trustworthy automation despite hyper-complexity~\cite{Mousavi:2020, ADF_C2:2018, UK_DOD_C2:2017}. That superior sense-making and learning, and by extension, swift and superior decision-making is achievable through accurate and adaptive mathematical functions to replace each stage in the OODA Loop~\cite{Kania:2017, Kalloniatis:2020, Wallace:2018, Wang:2020, Li:2018}. The risk entailed in doing so, will be discussed below. 

\subsection{The Risk of Fragility}

AI comes with new forms of risk that need to be managed. The phenomenon most consequential to a C2 system is the onset of war. If the outbreak of conventional state-on-state conflict (a very rare event) is missed, it could lead to a catastrophic surprise attack. Indeed, the opponent will be actively seeking a strategy to produce as large a shock as possible to the C2 system~\cite{Albino:2016}. The question that arises in this context is whether or not the benefits of automating C2 decision-making through AI algorithms is worth the risk of catastrophic failure? Is C2 performance improved holistically if it is prepared to automate decisions to deliver deadly force (or not) with an AI prediction of \textit{99\% confidence}, and the 1\% chance likely resulting in irreversible strategic consequences? For C2, the ramifications from getting strategic decisions wrong can be so extreme that it can lead to its own destruction, requiring an antifragile strategy as a necessity for survival against Black Swan events.

The reason an AI prediction with 99\% confidence can lead to failure, is due to the fact that AI suffers from what is called the \textquoteleft Platonic Fold\textquoteright \ when confronted with dynamic complex systems. The Platonic Fold describes the situation when a models \textquoteleft topology\textquoteright \ or \textquoteleft state space\textquoteright \ of a complex environment is inherently misinformed, or brittle, due to the details omitted \textquoteleft for the sake of hiding complexity\textquoteright~\cite{Taleb:2012, OReilly:2019, Antony:2016, Wallace:2018}. When complexity is hidden unwisely, the level of abstraction that the AI is operating with is simpler than the appropriate level of abstraction it should operate on. The result is emergent phenomena not represented in the AI state space or an inability to discriminate between different contexts requiring different decisions. These variables can be hidden reinforcing feedback loops that can lead to Black Swans, with often a catastrophic impact~\cite{Taleb:2012, Taleb:2020, Bruijn:2018, Davis:2018, Wallace:2018}. This poses risks for automated decision-making within the C2 operational environment. Worse still, even if an AI model is learning from the environment, it will become fragile if it cannot \textquoteleft keep up\textquoteright \ with changes in topology, developing more hidden variables as a result of time~\cite{OReilly:2019, Florio:2014}. Models that ignore or underestimate the impact of this uncertainty as an ontological fact of the complex environments they are trying to emulate, will produce increasing levels of fragility congruent to the consequence from model failure~\cite{Taleb:2012, Derbyshire:2013, Wallace:2018}.

Rapidly updating a model is planned to prevent the \textquoteleft drift\textquoteright \ associated with AI understanding of an \textquoteleft open\textquoteright \ and complex system. Florio~\cite{Florio:2014} argues that with regular training updates and enough unique data for training, a very complex model/function can be maintained to approximate a nonlinear system over time. This approach, often called the \textquoteleft ML pipeline\textquoteright \ or ML Development Process~\cite{Allen:2020}, is a cyclical technique in which one ML model is operational and predicting the environment, while another is being trained. Changes in the environment only lead to new data for the algorithm to update itself on, improving the repository of models for the C2 system to draw on as it orients its activities to the environment. The rate in which a model is updated and replaced will have a corresponding impact to the \textit{fidelity} of the model in accurately reflecting a complex environment~\cite{Florio:2014}.

However, rapid model updates does not solve the Platonic Fold for a decision-making AI. An ML model can be rapidly updating a continuously inaccurate model, and totally unaware of degradation in data~\cite{Wallace:2018}. AI could quickly result in a C2 system with optimised and superior decision-making for events it has been trained on, at the expense of being brittle or fragile to events that have yet to occur or be sensed by the system~\cite{Wallace:2018}. However, as discussed above, it is precisely these rare events that have yet to occur that the C2 system considers its highest priority.

The point of system failure for AI-enabled C2 is when the AI model makes rapid decisions that aid in the collapse of control, resulting in helplessness or confusion, due to the mismatch between the topology of the operational environment and the representational topology~\cite{Wallace:2018, Albino:2016}. As an example, Wallace~\cite{Wallace:2018} discusses recent \textquoteleft flash-crashes\textquoteright \ (Black Swans) of the stock market as analogous to the results that should be expected from fragile AI in a C2 system. These crashes occurred due to automated trading algorithms that were too rapid for human intervention, with underlying causes so complex that they are still unknown. For C2, the equivalent could be two opposing military\textquoteright s with highly autonomous AI decision-making, resulting in a flash-crash of high intensity war; all from a loss of stability measured in milliseconds~\cite{Wallace:2018}.

\subsection{C2SIM and AI}

The proposed solution to the risk of AI missing rare and catastrophic events is through the use of synthetic (artificially constructed) data. Synthetic data is the only realistic method enabling an ML algorithm to train from data on phenomenon of high interest to a C2 system, such as the conventional high-intensity wars of the future that the C2 system is designed to effectively make decisions in~\cite{Jin:2018, Wang:2020, Mousavi:2020}. No data exists for the wars of the future, and it is arguable whether the wars of the past would be useful. The process of synthetic data generation falls under three categories~\cite{Newlin:2019}: 
\begin{enumerate}
    \item Manual development, through curated datasets built by hand.
    \item The automatic tweaking of real inputs to generate similar inputs to help the algorithm learn broader rules.
    \item Automatically through modelling and simulation (M\&S) and emulation.
\end{enumerate}
Which process to use depends entirely on the purpose of the AI and the scarcity of the environment that it is trying to make predictions from. If the AI is to replace the decision-making capabilities of a commander, it is highly likely that a combination of manually created data derived from intelligence, alongside simulated models of the battlefield, will be required to train an AI-enabled system. This method integrates concepts such as C2SIM and AI together, possibly with the use of reinforcement learning algorithms~\cite{Mousavi:2020, Zhang:2020}. 

However, risks persist in this approach. Creating a highly detailed model of the operational environment is not only very difficult to validate, but would likely produce results that are deceiving, as the AI would lack the fidelity required to make effective decisions under uncertainty~\cite{Davis:2018, Mousavi:2020, Zhang:2020}. Davis~\cite{Davis:2018}, describes this as a reduction in the \textquoteleft scenario-space\textquoteright, meaning that the options or flexibility that the AI is trained on becomes narrow. An AI system developing courses of action for a commander within a C2 system that is optimised to specific scenarios, will only have reliable performance as a reactive system to a highly specific scenario-space. The assumption of causation, or non-causation, between variables within the model will inevitably lead to fragility \cite{Davis:2018}. 

On the other hand, a highly abstracted model that ignores most of the finer details of the operational environment for a more \textquoteleft strategic level\textquoteright \ recommendation system, has its own problems. The use of synthetic data will be inseparable from the military culture that is creating it. The assumptions made about the enemy and how they will fight the next war, will be cemented in the data that the AI is trained on~\cite{Wallace:2018}. If an enemy decides to \textquoteleft change the rules of the game\textquoteright, with an asymmetric action at the strategic level that the AI has never been trained on, any novel enemy strategies or tactics will not be accurately predicted on the outset of the occurrence~\cite{Zhang:2020}. Instead, they will be predicted as something else entirely. At the strategic level, such as the theatre of battle, the variables associated with predicting enemy behavior will have long statistical \textquoteleft tails\textquoteright \ not represented in the AI model~\cite{Wallace:2018}. this could have serious strategic consequences, leading to a system not suited for the \textquoteleft deep uncertainties\textquoteright \ or volatility of war~\cite{Davis:2018, Zhang:2020}. Zhang~\cite{Zhang:2020} noted that the use of AI \textquoteleft for applications involving strategic decision-making, such as those where simulations do not even have physics to fall back on, may have so little correspondence between the real world and the simulation, that trained algorithms will be effectively useless\textquoteright. It follows that for AI to remain useful, it must be trained from data that corresponds to a C2 function that is adequately complicated - not complex. Clearly, in order for AI to be used without becoming a fragility risk, a balance needs to be struck between trust in the AI, the risk of prediction failure, and the benefits of responsiveness the specific AI brings to a C2 function.

\begin{figure}[ht]
\centering
\includegraphics[scale=0.3]{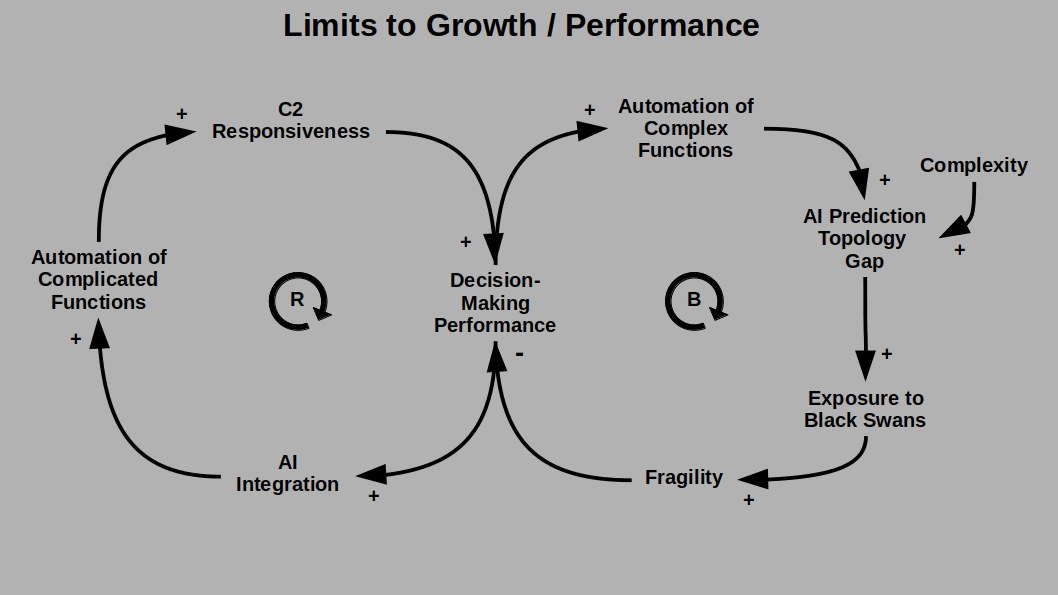}
\caption{AI Integration and the Limits to Growth}\label{fig4}
\end{figure}

The risk of fragility associated with AI-enabled C2 systems, reflects the archetype of limits to growth, displayed above in Figure~\ref{fig4}. Decision-making performance is improved through the automation of complicated functions, resulting in increased C2 responsiveness. However, AI integration into more complex functions (such as decision-making) results in more risk being transferred to the accuracy and variance of the AI model compared to the operational environment. This could then lead to prediction failures from low probability but high consequence catastrophic events. The more functions that the AI is replacing that need context and judgement to understand the complex environment, the more fragile the system will become. Black Swan events are both mathematically unpredictable and consequential to a system. Therefore, the more risk the C2 system is exposing itself to significant shocks, the more likely it is to eventually suffer catastrophic failure~\cite{Taleb:2012, Taleb:2020, Cirillo:2015, Bruijn:2018}. 

\section{From AI Fragility to Antifragility}\label{sec:Antifragility} 

The method for integrating AI into an Agile C2 system without increasing exposure to fragility will require careful consideration of the antifragile elements discussed above in Table~\ref{tab1}. Specifically, the C2 system will need to ensure a convex response to shocks from the operational environment. This can be achieved through two methods:
\begin{enumerate}
    \item Function allocation of AI into a C2 system to minimise risk from catastrophic failure but maximise gains to the system. 
    \item Innovativeness and chaos generation using experimentation to discover fragility in the system; this enables overcompensation and AI model variance. 
\end{enumerate}

\subsection{Function Allocation}

An AI-enabled system requires a balance between its implementation as a tool for agility and its potential to become a fragility risk if the AI does not perform under extreme volatility within a complex environment.  AI is not suitable for all decision-making tasks~\cite{Alberts:2020, Abbass:2019, Kalloniatis:2020}. An antifragile system will require specified boundaries to separate the C2 decision-making functions with high exposure to Black Swans at the strategic/operational level, from other complicated C2 functions that can be automated with low exposure. A clear articulation of what tasks AI will be responsible for within a C2 system will be crucial to avoiding fragility and benefiting the system holistically.

Due to the fact that a C2 system is sociotechnical, those that allocate the use of AI for C2 functions need to ensure that the replacement of a human does not risk the performance of the system. Abbass~\cite{Abbass:2019}, discusses several methodologies for the allocation of AI in such a system. A \textquoteleft static allocation\textquoteright, whereby functions are given to AI in the C2 system  and do not change, is likely not suitable for a dynamic environment. The needs of the specific C2 function will change depending on circumstances, especially when considering the need for responsiveness in warfare, which may require a rapid change in function allocation~\cite{Kalloniatis:2020}. For example, a hyper-sonic missile defence scenario against incoming volleys will prefer speed over the strategic context. The consequences of doing nothing in this scenario are so great, that the risk of being wrong \textit{may} be worth full AI control. On the other hand, a decision to approve a hyper-sonic attack will require more context for the decision than speed. An adaptive approach, or Automated Allocation Logic (AAL) is therefore necessary~\cite{Abbass:2019}.

\begin{figure*}[ht]
\centering
\includegraphics[scale=0.5]{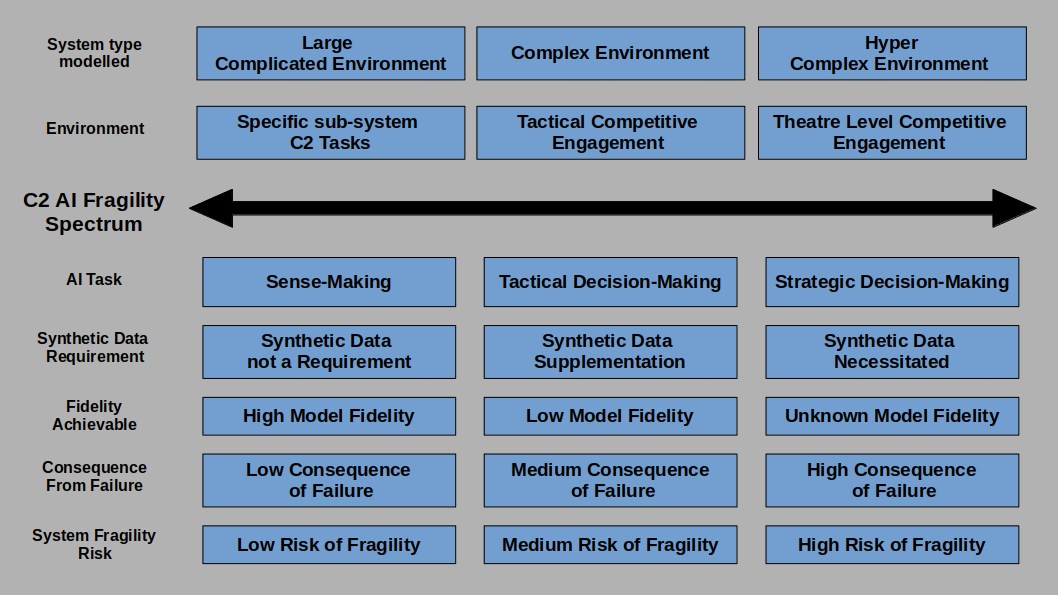}
\caption{AI-enabled C2 Fragility Spectrum}\label{fig5}
\end{figure*}

At the strategic decision-making level, a critical event logic is most appropriate for assessing the fragility against the benefits of automation. Depending on how critical the need is for responsiveness and how high or low the consequences are from failure, C2 functions will need to have adaptive logic for human or AI control. Figure~\ref{fig5} presents an example for the potential consequences associated with broad categories of C2 tasks, from sense-making to theatre-level decision making. 

For an AI system that focuses on sense-making AI, the risk is lower, due to the extra context applied to the data from human decision-makers~\cite{Kalloniatis:2020}. Sense-making will likely require multiple specialised algorithms to parse specific categories of data, such as video feeds, pictures, documents and others~\cite{Allen:2020}. It is therefore also a robust system of algorithms, if one algorithm fails to sense critical information, it is less likely to be completely missed by all other sensors. Of course, risk still persists, and this will need to be assessed through an understanding that a \textquoteleft transference of risk\textquoteright \ in decision-making has been passed to the inputs and sense-making capabilities of the AI system~\cite{Abbass:2019}. 

AI Decision-making, however, as discussed above, is associated with higher exposure to the risks of failure during war. The impact of failure will depend on whether the AI is supporting the tactical level, operational level, or strategic; a single failure on the tactical level will have less consequences compared to a single failure on the strategic level; although, one must account for the possible cascading of effects from tactical to strategic levels. For an antifragile system, Taleb~\cite{Taleb:2012} states that one should avoid the reliance on systems with highly consequential output, as many smaller, less consequential systems are less fragile.  Of course, even if the risk from a strategic level AI decision-maker is managed through a human-in-the-loop construct, risk still remains due to the recommendations relying on AI sense-makers, and the additional effect of forecasting on human decision-makers. For example, if a C2 system uses a trusted Non-human Intelligence Collaborator (NIC) to recommend decisions at the strategic level, it may lead to an increase in risk taking by military commanders who are presented the 99\% confident AI prediction. This is because the NIC will behave as a forecaster, and evidence indicates that this could increase risk taking amongst decision-makers~\cite{Taleb:2012, Shrader:2020, Tombu:2015, Albino:2016}. 

Once the consequence of failure has been determined, the adaptive AI for each scenario will then need to be assigned. This is a \textquoteleft command concept\textquoteright \ C2 function; the commanders intent and the nations strategic objectives will need to be considered when an adaptive AI function is allocated for specific scenarios. These scenarios can be developed and tested through the traditional method of war gaming, but can also emerge from the antifragile process of innovativeness and chaos generation. adaptive AI will need to be consistently tested for fragility in order to prevent a concave response; the subject of the next section.  

\subsection{Innovativeness and Chaos Generation}

In order to implement AI as an agile and antifragile tool, the elements of feedback/memory, small scale experimentation and overcompensation, need to be combined in an AI-enabled C2 system construct. This can be done through purposely injecting volatility into the system, and by extension, the AI functions supporting specific C2 processes. Through the use of volatility, an AI system will develop a broader/abstracted decision-space, increasing its versatility to a wider variety of shocks.

For synthetic data generation, a consistent degree of volatility and chaos could be applied to data the AI is trained on. For example, extreme scenarios could be tested on the AI system instead of just \textit{expected} extreme scenarios. A \textquoteleft chaos team\textquoteright \ within the C2 organisation could seek to try and expose the prediction failures in AI models, using extreme or highly unlikely scenarios. Through exposing failure, the AI development team could then determine why the failure took place, an exploration on what action by the AI would have been preferable, and then an attempt to retrain the model to increase its variance to handle a similar extreme scenario in the future. This process thus strengthens the system via an understanding of itself compared to the complex environment outside it~\cite{Taleb:2012}. It is possible that this could also be achieved by an AI scenario generator, with a primary purpose to be rewarded for developing scenarios that lead to failure for the AI-enabled C2 system. No matter the exact method, the aim is for system stress and failures to allow the innovativeness and discovery within the C2 system to occur, resulting in overcompensation.

These shocks are not just required for AI itself, but for the C2 system holistically. A layered approach, as a form of robustness, should be sought~\cite{Taleb:2012}. One method for doing so can be found in computational red teaming and Chaos Engineering practice. Computational red teaming~\cite{abbass2016computational} offers the computational building blocks required by an AI to design stressors to challenge itself and the environment it is situated within, and evolve new models and tactics. In a similar manner, Chaos Engineering prevents fragility within an organisation through experimentation with injected stress to, or deliberate failure of, specific elements in a computer network or system~\cite{Rosenthal:2017}. The aim of chaos engineering is to ensure \textquoteleft availability\textquoteright \ of all the functions of the C2 IT system despite volatility in the environment. The usefulness for antifragile C2 is obvious in that its Chaos Engineering experiments allow for the generation of operational environment effects, such as cyber attacks, as inputs for extreme volatility. The C2 IT and communications network is viewed as a single complex system, which is better understood through observing its behaviour after real-world inputs or induced failures~\cite{Rosenthal:2017}.

\begin{figure}[ht]
\centering
\includegraphics[scale=0.3]{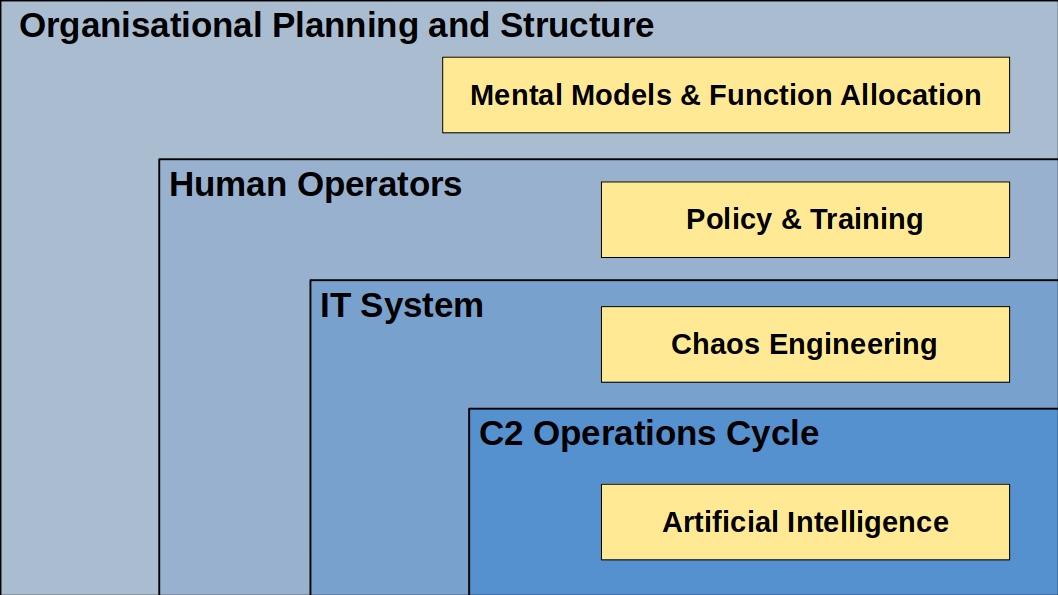}
\caption{Antifragile C2 as a System of Systems}\label{fig6}
\end{figure}

Combining Chaos Engineering, computational red teaming and AI can enable a sophisticated generation of failure states to enable antifragility, but a large change in organisational culture is required for a C2 system to have the ability to learn from self inflicted stress in order to overcompensate. Seen in Figure~\ref{fig6} is the A3IC2 system of systems. Creating such a system of systems within a C2 organisation will necessitate a shift in organisational mental models, organisational planning, C2 structure, and a change in how human operators are trained for supporting an antifragile C2 system. A3IC2 should be only concerned with the system of C2 operations; the process of conducting C2 successfully as an antifragile system. For a C2 organisation to be completely antifragile as a sociotechincal system, it will need to take a holistic approach, with structures, systems, processes and cultures all taking on antifragile qualities in order to survive stresses and shocks~\cite{Kennon:2015}. 

\begin{figure*}[ht]
\centering
\includegraphics[scale=0.6]{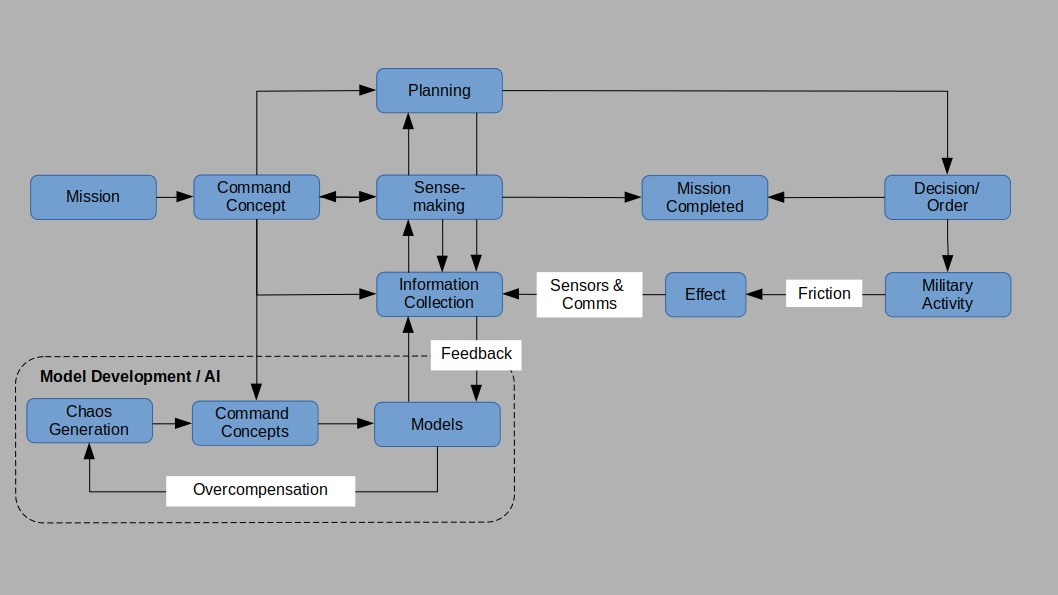}
\caption{Antifragile Dynamic OODA Loop}\label{fig7}
\end{figure*}

\section{Agile Antifragile Command and Control (A3IC2)}\label{sec:A3IC2} 

Through incorporating the antifragility concept with the functional C2 models developed by Boyd, Brehmer and Alberts~\cite{Brehmer:2005, Osinga:2007, Alberts:2011}, a new framework for improving the effectiveness of C2 systems through antifragility dynamics can be developed. This is seen below in Figure~\ref{fig7}, illustrating the difference between the traditional C2 operations cycle in Figure~\ref{fig1} and the A3IC2 structure.

Figure~\ref{fig7} describes the same DOODA Loop created by Brehmer, with the addition of feedback for the accumulation of models implemented. The creation of models serves as the systems method for learning from interacting with the complex environment during operations. The amalgamation of feedback from decisions made, planning, sense-making activities, and the results of military action, all provide the context for AI model/function. The models developed are dependent on the specific C2 system. For an air mobility/logistics C2 unit, the model would reflect decisions such as priority, aircraft selection, routes chosen, and cargo validation details, amongst others. For a AI-enabled C2 recommendation system for COA development, the feedback would represent variables such as enemy location, blue location, number of units, amongst many others. These models are built through interaction with the C2 Decision Support System during daily operations and/or through C2SIM.

As discussed above, the \textquoteleft chaos generation\textquoteright \ function is the method that forces an overcompensation from what the system has learned from feedback. It applies to both the human and machine within the sociotechnical system. Chaos generation is the C2 \textquoteleft red team\textquoteright \ that purposely stresses the system in order to strengthen the decision-cycle and improve agility and reduces fragility. For a AI-enabled C2 system, the chaos generator includes the synthetic data generation process based on prior experience, but modifies it to stress the system. The AI will therefore be trained and improved on missions with more extreme variables beyond previous experience; resulting in overcompensation. The models may be extreme in nature and should cover as much of the possibility space as it can. If a significant change in the environment occurs, or a Black Swan, the possibility space only increases, allowing for the system to improve and generate further models. The more volatility to the C2 system, the more models are produced to compensate. 

Previous discussion assumes that models and data need to be built in advance, and in anticipation of the future to come. Recent trends have introduced models that get formed, reshaped, and calibrated in-situ. The Shadow Machine concept~\cite{abbass2016computational} has a dedicated control logic to learn the model as the context unfold. However, these concepts assume a real-time datafeed from the actual context to continuously measure deviations and adapt accordingly. Challenges still remain with this approach. Data about self is likely to be orders of magnitude more than data about the enemy. This imbalance in the data available for the AI to learn the models on-the-fly has its own challenges within the AI community.

\section{Conclusion}\label{sec:conclusion}

The integration of AI into C2 will only improve the performance of the system if it is implemented through holistic understanding of its effects. If an AI-enabled C2 function has the possibility to contribute to the failure in delivering the strategic objectives of the nation it is defending, then serious consideration needs to be made about the efficacy of that AI. When C2 functions are allocated to AI to avoid fragility, then the use of feedback and overcompensation has the potential to facilitate a convex response to system volatility. The use of purposeful chaos generation will aid the C2 system to enable knowledge of its own weaknesses in order to improve. The use of A3IC2 as a strategy for AI-enabled C2, can ensure that AI remains as a tool towards building an antifragile system. Minimising the potential for catastrophic failure, while maximising the exploitation of benefits to the system, will enable both survival and victory during the extreme volatility of war. 

While the focus of this paper has been on the risks faced by an AI, a human commander will still be faced with similar issues when novel situations unfold, especially when lessons from military history could hinder their ability to think about these new situations. Future conflict scenarios will be more challenging if the enemy is relying on AI to generate effects near the speed-of-light. This calls for a human-AI teaming approach to leverage the strength of each and overcompensate for their individual weaknesses to generate effects at the speed of relevance.

\bibliography{references}
\bibliographystyle{IEEEtran}

\end{document}